\documentclass[10pt,twocolumn,letterpaper]{article}

\usepackage{cvpr}              % To produce the CAMERA-READY version
\usepackage{graphicx}
\usepackage{amsmath}
\usepackage{amssymb}
\usepackage{booktabs}
\usepackage{multirow}
\usepackage{bm}

\usepackage[pagebackref,breaklinks,colorlinks]{hyperref}

\usepackage[capitalize]{cleveref}
\crefname{section}{Sec.}{Secs.}
\Crefname{section}{Section}{Sections}
\Crefname{table}{Table}{Tables}
\crefname{table}{Tab.}{Tabs.}

\def\cvprPaperID{*****} % *** Enter the CVPR Paper ID here
\def\confName{CVPR}
\def\confYear{2022}

\begin{document}

% \title{CFSmoke: A coarse-to-fine method for detecting smoking cigarettes}
\title{Application-Driven AI Paradigm for Hand-Held Action Detection}
\author{Kohou Wang\\
ChinaUnicom\\
{\tt\small wangzp103@chinaunicom.cn}
\and
Zhaoxiang Liu \thanks{Corresponding author} \\
ChinaUnicom\\
{\tt\small liuzx178@chinaunicom.cn}
\and
% Shiguo Lian \thanks{Corresponding author} \\
Shiguo Lian$\ ^*$ \\
ChinaUnicom\\
{\tt\small liansg@chinaunicom.cn}
}

% \author{Shiguo Lian \\
% ChinaUnicom\\
% {\tt\small liansg@chinaunicom.cn}
% }
% % \thanks{Corresponding author: email@mail.com}}
%
% \author{Zhaoxiang Liu \\
% ChinaUnicom\\
% {\tt\small liuzx178@chinaunicom.cn}
% % \thanks{Corresponding author}
% }
% \authornote{*Corresponding authors}
% \cortext[cor1]{Corresponding author}
\maketitle

% \author{first author}
% \author{ShiGuo Lian\corref{mycorrespondingauthor}}
% \author{ZhaoXiang Liu\corref{mycorrespondingauthor}}
% \cortext[mycorrespondingauthor]{Corresponding author}
% \ead{\tt\small firstauthor@i1.org}
% \maketitle
% \address{School of Mathematical Sciences, University of Science and Technology of China}
%\author{Third Author}
%\address{Department of Computer Science, Tel Aviv University}

%%%%%%%%% ABSTRACT
\begin{abstract}
    In practical applications especially with safety requirement, some hand-held actions need to be monitored closely, including smoking cigarettes, dialing, eating, etc. Taking smoking cigarettes as example, existing smoke detection algorithms usually detect the cigarette or cigarette with hand as the target object only, which leads to low accuracy. In this paper, we propose an application-driven AI paradigm for hand-held action detection based on hierarchical object detection. It is a coarse-to-fine hierarchical detection framework composed of two modules. The first one is a coarse detection module with the human pose consisting of the whole hand, cigarette and head as target object. The followed second one is a fine detection module with the fingers holding cigarette, mouth area and the whole cigarette as target. Some experiments are done with the dataset collected from real-world scenearios, and the results show that the proposed framework achieve higher detection rate with good adaptation and robustness in complex environments.
    % 1.举例：加油站、化工厂、工厂车间；
    % 2.室内安全事故、爆炸、吸烟有害人体健康；吸烟检测非常有价值；
\end{abstract}
\section{Introduction}
\label{sec:intro}
In practical applications especially in indurstial scenarios, some hand-held human actions need to be monitored closely, including smoking cigarettes, dialing, eating, etc. Under most circumstances, smoking cigarettes is an important problem of safety. Places like chemical plants, industrial workshops and petrol stations, all require strict control of smoking cigarettes, which may bring about fire, explosion, etc. For instance, a spark in some chemical plants may lead to a disaster of a lot of human life. Additionally, in most scenarios, it is not permitted to use smart phone or eating something during on-duty. In this paper, we take smoking cigarettes as a typical example to discuss how to detect this kind of hand-held action.
\begin{figure}[t]
  \centering
    \includegraphics[width=1\linewidth]{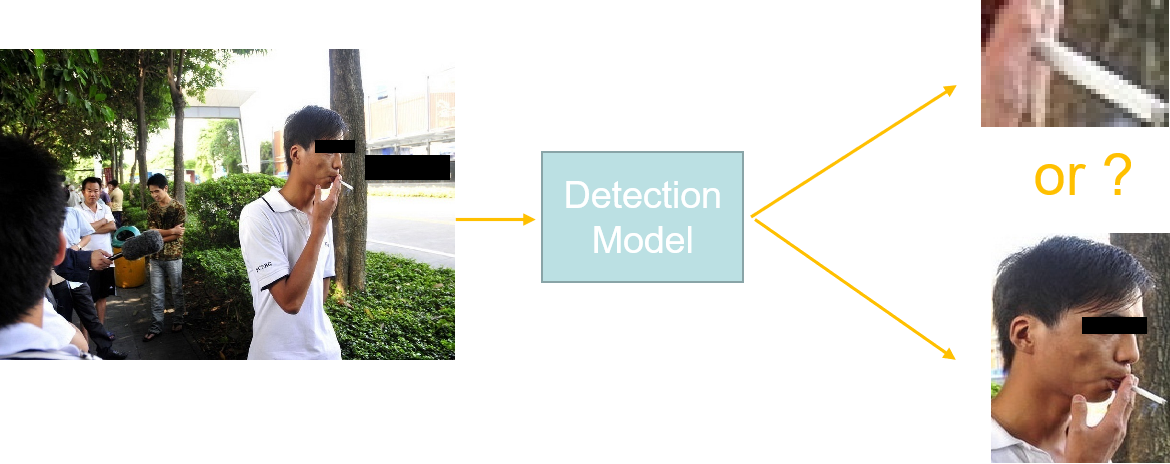}
  \caption{How to detect smoking cigarettes in real-world scenarios? Generally, one may focus more on the cigarette itself while the other on the human smoking pose.}
  \label{fig:mainstream methods flown}
\end{figure}
\begin{figure}[t]
  \centering
    \includegraphics[width=0.8\linewidth]{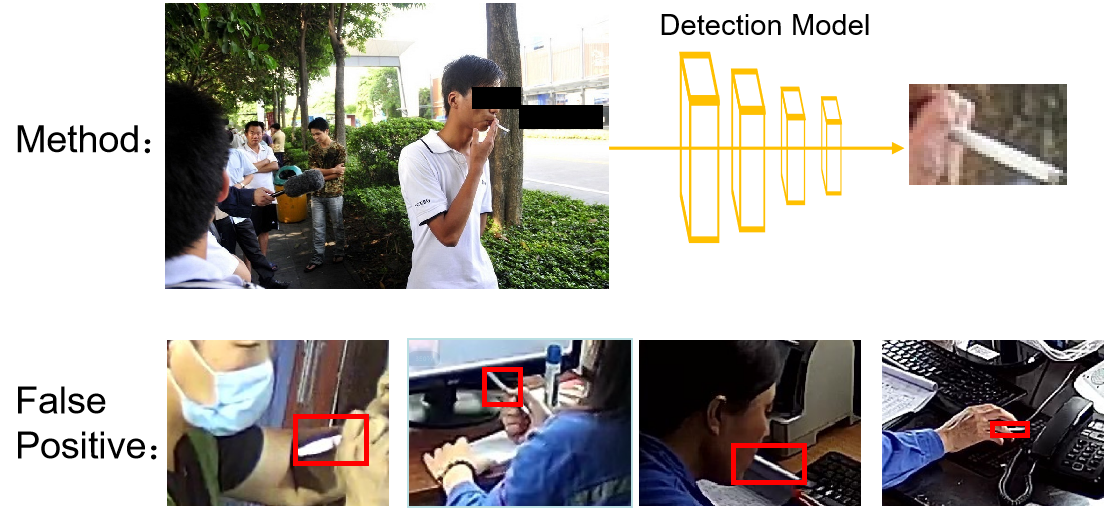} %插入图片，[]中设置图片大小，{}中是图片文件
  \caption{Method focusing only on the cigarette itself is more inclined to falsely take sticks that are similar to cigarettes as targets.}
  \label{fig:finer methods}
\end{figure}
\begin{figure}[t]
  \centering
    \includegraphics[width=0.8\linewidth]{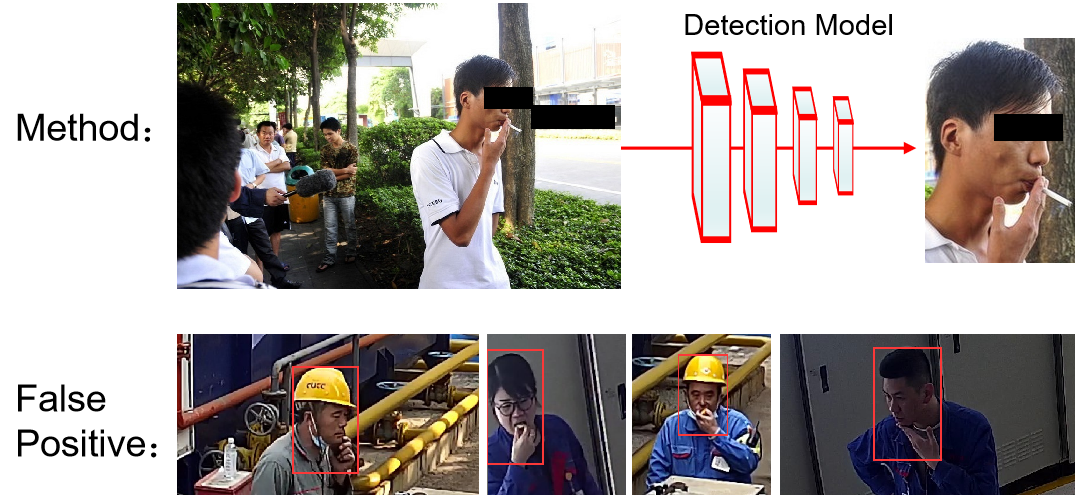} %插入图片，[]中设置图片大小，{}中是图片文件
    % \label{fig:short-a}
  \caption{Method focusing on the human smoking pose may overlook the cigarettes and take only the body pose as target.}
  \label{fig:overall methods}
\end{figure}
\begin{figure*}
  \centering
    \includegraphics[width=1\textwidth]{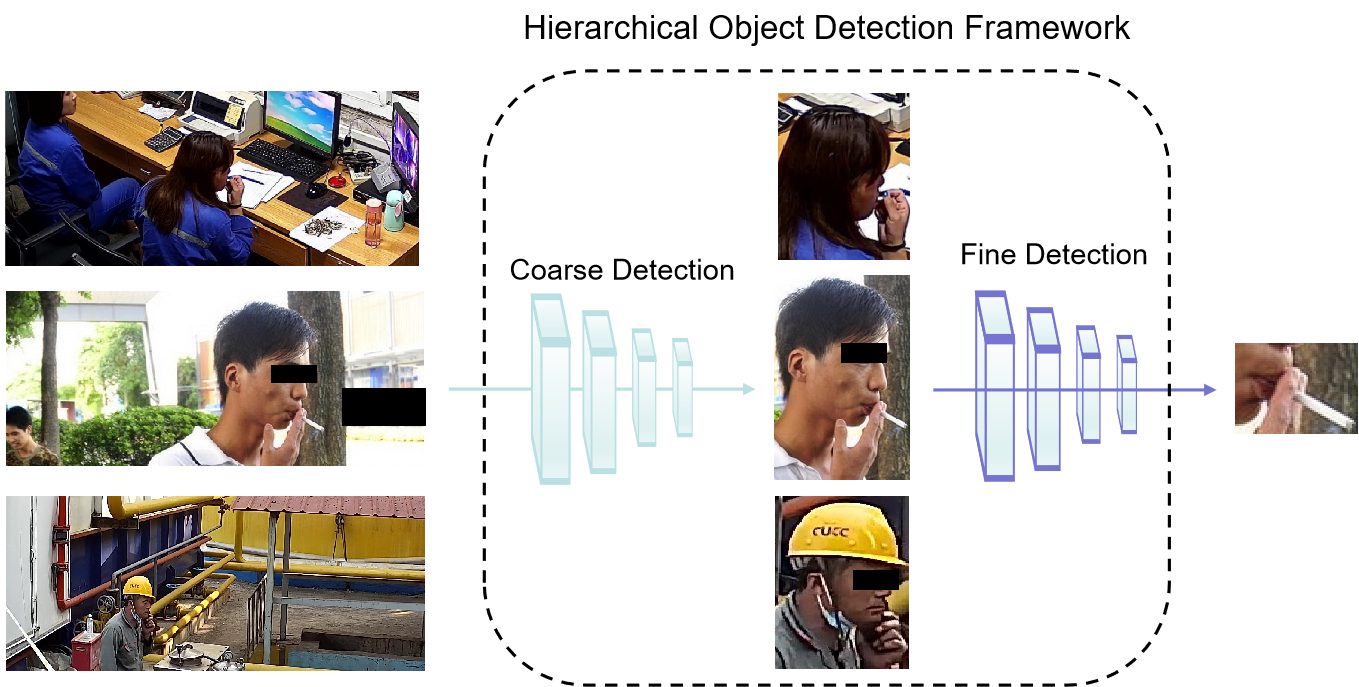} %插入图片，[]中设置图片大小，{}中是图片文件
    % \label{fig:short-a}
    %\caption{Mainstream methods of detecting smoking cigarettes can be roughly devided into two kinds: one kind of method focuses on the cigarettes itself while the other kind focuses on the overall human body and smoking pose.}
  \caption{Our hierarchical object detection framework composed of two detection modules. The first module focuses on a coarse object consisting of the whole hand, cigarette and head. The followed second module detects the object consisting of the fingers holding cigarette, mouth area and the whole cigarette.}
  \label{fig:method-flow}
\end{figure*}
With the rapid development of deep learning\cite{schmidhuber2015deep}, more and more detection methods are utilized in the field of smoking cigarettes detection\cite{ortis2020report}. However, most of these methods adopt only one single detection model, which may conquer one kind of problem but meanwhile lead to another problem. In other words, due to the inherent attributes of the target object, methods utilizing one single model are scarcely possible to cover all kinds of problems that may occur under real-world scenarios.

Intuitively, methods of detecting smoking cigarettes can be roughly devided into two kinds: one kind of method focuses on the cigarettes itself while the other focuses on the overall human smoking pose, as shown in Figure \ref{fig:mainstream methods flown}. Typical distribution of real-wolrd scenarios' smoking cigarettes can be roughly classified into three classes: (a) images having no cigarettes but human body pose are similar to smoking cigarettes; (b) images having cigarettes and are easy to be detected correctly; (c) images having sticks similar to cigarettes but having no cigarettes. As for these different kinds of scenarios, current mainstream methods can hardly cover all these cases and achieve a high accuracy. This is because the class (a) and (c) are exactly two opposite directions for training models: a model focusing its attention on class (a) is natively inclined to neglect the small cigarettes compared to the much bigger human body, and vice versa. For instance, as is shown in Figure \ref{fig:finer methods}, if one method focuses more attention on the cigarette itself, it's more inclined to falsely take sticks that are similar to cigarettes as the final target; but also,  as is shown in Figure \ref{fig:overall methods}, if one method focuses its attention on the overall human body and smoking pose so as to amend the former stick mistakes, it again may bring about other problems: it may overlook the cigarettes and take the bigger pose and human body as final target instead.

To adress these issues, the hierarchical coarse-to-fine detection framework is proposed in this paper as a new application-driven AI paradigm. In this framework, the coarse detection module detects the target of human smoking pose consisting of the whole hand, cigarette and head, while the followed fine detection module detects the target of cigarette consisting of the fingers holding cigarette, mouth eara and the whole cigarette. With the hierarchical framework, both the overall pose and details are considered, which confirms the high accuracy. The rest of the paper is arranged as follows. The related work is introduced in Section 2. In Section 3, the proposed application-driven AI paradigm is presented in detail. The experiments are done and results are given in Section 4. Finally, in Section 5, the conclusion is drawn.

\section{Related Work}
Some methods have been proposed to handle the problem of detecting smoking cigarettes. The method  \cite{zhang2018smoking} proposes a smoking image detection model based on a convolutional neural network, referred to as SmokingNet, which automatically detects smoking behaviors in video content through images, this method can detect smoking images by utilizing only the information of human smoking gestures and cigarette image characteristics without requiring the detection of cigarette smoking. The method \cite{senyurek2020cnn} proposes a novel algorithm for automatic detection of puffs in smoking episodes by using a combination of Respiratory Inductance Plethysmography and Inertial Measurement Unit sensors. The detection of puffs was performed by using a deep network containing convolutional and recurrent neural networks. The method \cite{ruilong2021algorithm} designs a series of convolutional neural network modules to reduce the amount of model parameters and pick up the inference speed to meet real-time requirements as well as improving the accuracy of small target object (cigarette) detection.

Generally, in these methods, an object detection model is adopted to localize the smoking action in an image. In nowadays, the object detection models constructed on CNNs are popularly used, such as YOLOv5\cite{glenn_jocher_2022_6222936}, faster rcnn\cite{ren2015faster-rcnn}, retinanet\cite{lin2017focal}, cornernet\cite{law2018cornernet}. For the video or temporal image sequence composed of consecutive image frames, the object detection model may be applied to each image to give the detection result, and the combination of consecutive images' results gives the final result of the video or temporal image sequence.

The mainstream approach for smoke detection can be divided into two kinds\cite{chien2020deep, ruilong2021algorithm, shi2021faster, han2019cigarette}: one method, simple and easy to think of, is to detect the smoke itself, ignoring all kinds of context information, which simplifies this method's processing procedure but meanwhile leads to more wrong detections; the other method, which takes care of more context and specific scenarios, involves in more other human bodies, including hand, mouth, upper body, body pose, etc. However, these two methods, having their own advantages, both have some certain intrinsic vulnerablenesses. The first method may take all kinds of sticks, which are similar to cigarettes, as cigarettes by mistake. The second method, which to a certain degree decreases the first method's mistakes, may take some poses similar to smoking cigarettes as target by mistake. In this paper, the two kinds of methods are combined in order to avoid their individual disadvantages.

 \begin{figure}[t]
   \centering
     \includegraphics[width=0.8\linewidth]{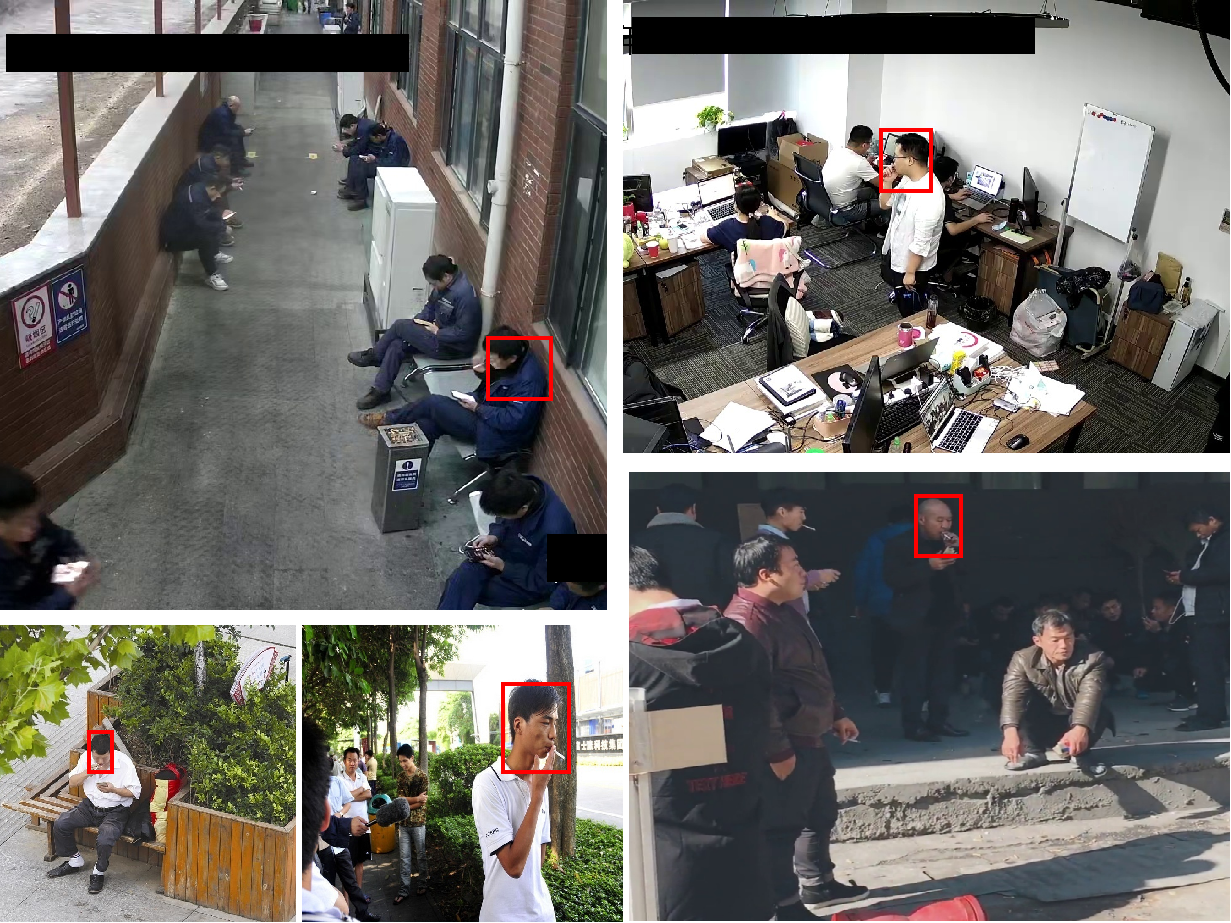} %插入图片，[]中设置图片大小，{}中是图片文件
     % \label{fig:short-a}
   \caption{Examples of the coarse detection model's annotation. The target object consists of the whole hand, cigarette and head.}
   \label{fig:roughModle}
 \end{figure}
 \begin{figure}[t]
   \centering
     \includegraphics[width=0.8\linewidth]{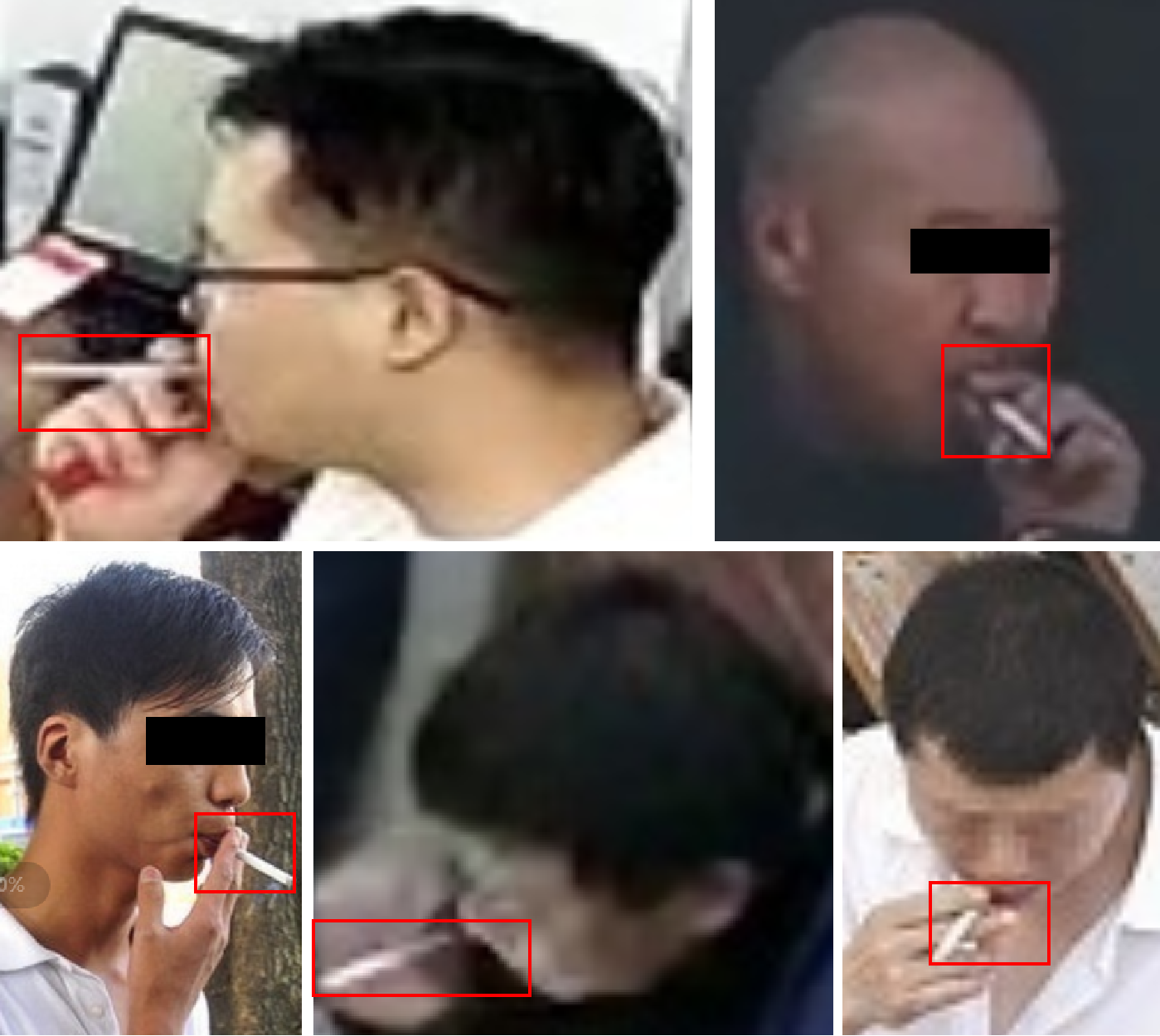} %插入图片，[]中设置图片大小，{}中是图片文件
     % \label{fig:short-a}
   \caption{Examples of the fine detection model's annotation. The target object consists of the fingers holding cigarette, mouth area and the whole cigarette.}
   \label{fig:finerModle}
 \end{figure}
 \begin{figure*}
   \centering
   \begin{minipage}{0.9\linewidth}
     \centering
      \includegraphics[width=0.8\linewidth]{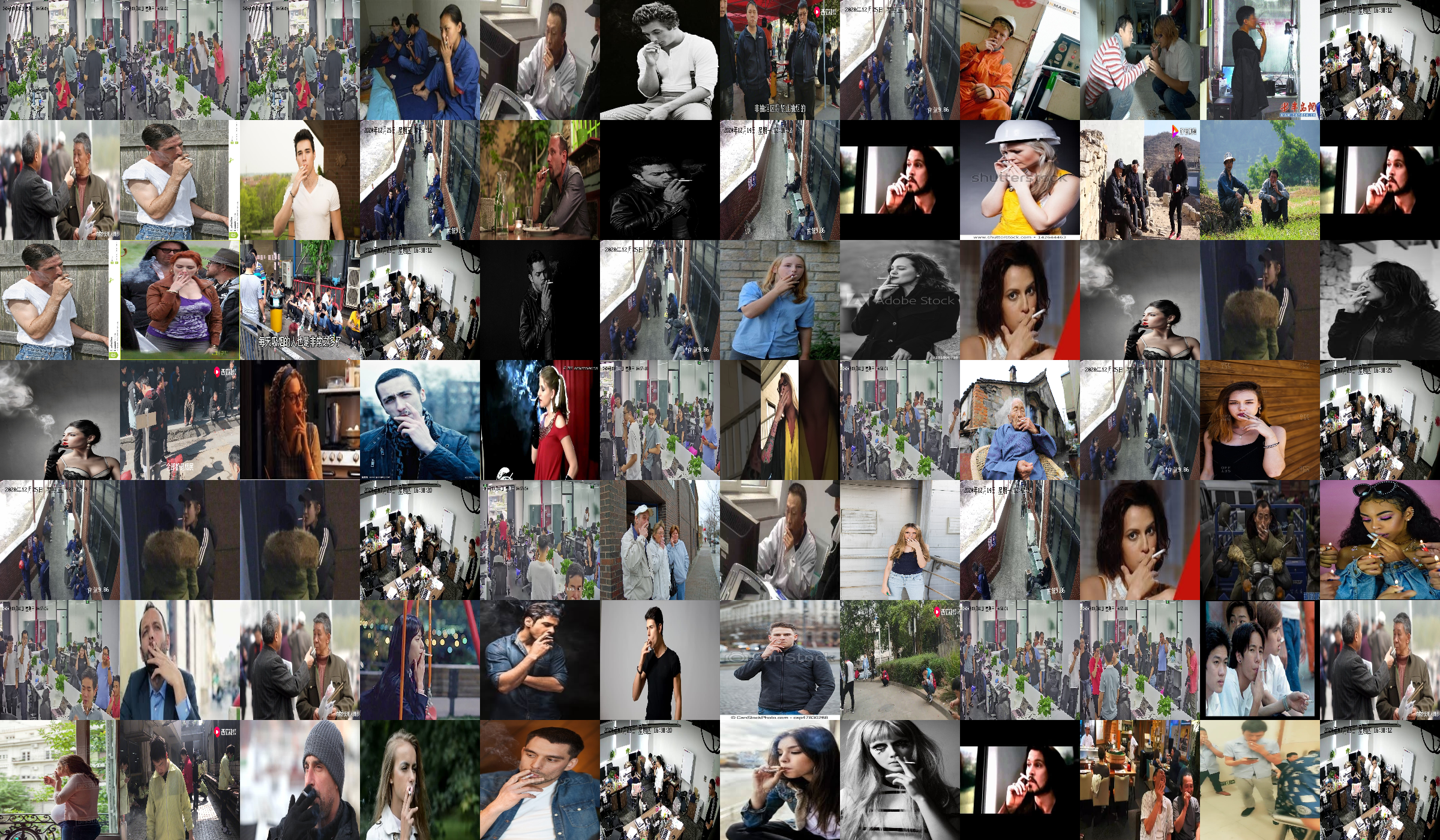}
      \caption{A glimpse of our coarse detection model's dataset. The target object consists of the whole hand, cigarette and head.}
      \label{fig:original-images}
   \end{minipage}
    \begin{minipage}{0.9\linewidth}
      \centering
       \includegraphics[width=0.8\linewidth]{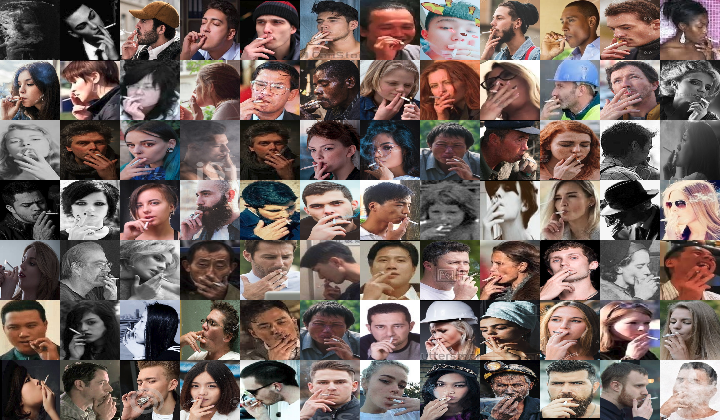}
       \caption{A glimpse of our fine detection model's dataset. The target object consists of the fingers holding cigarette, mouth area and the whole cigarette.}
       \label{fig:normalbbox}
     \end{minipage}
 \end{figure*}
\section{The Proposed Application-Driven AI Paradigm}
\label{sec: SYSTEM OVERVIEW}
\subsection{The hierarchical object detection framework}
In this paper, we propose a coarse-to-fine two-stage method to deal with this dilemma. Our method utilizes two detection modules: the first module focuses on a coarse object of human smoking pose, while the followed second one focuses on the fine object of the cigarette. Overview of the two stages is shown as Figure \ref{fig:method-flow} and detailed as follows.

\subsection{Coarse detection module}
The coarse detection module roughly detects an object including whole hand, cigarette and head, as shown in Figure \ref{fig:roughModle}. In this detection, we take a typical human pose of smoking cigarettes as target. A pose of smoking consists of several parts, but the typical and essential one is to hold a cigarette and feed it into mouth, which we take as the target of the first stage's detection module. In this way, hands and whole head, as context of a typical pose of smoking cigarettes in an image, can filter out most targets that can be easily misdetected as smoking cigarettes.

\subsection{Fine detection module}
The second module detects a finer object consisting of the fingers holding cigarette, mouth area and the whole cigarette. Following the coarse detection module, the second module detects a much smaller object as shown in Figure \ref{fig:finerModle}. In this way, the second module's results, which are based on the first module's input, can natively utilize the first module's design philosophy to filter out a majority of mistakes, meanwhile the well designed finer object can filter out other kinds of false detections: for instance, due to the relatively bigger target, the first module may improperly take hands and whole head as the target of smoking cigarettes and ignore the cigarettes. This is because in the bigger target, the cigarette itself is a much smaller object compared to hands and human head, after several CNN layers, the feature map may just take the cigarette as an irrelevant noise of the overall target. Hence the second module concentrates more on a rather small target: the fingers holding cigarette, mouth area and the whole cigarette. In this smaller area, the module will give more attention to the cigarettes, while still give consideration to other body parts of fingers and mouth, which in all are pivotal components of a pose of smoking cigarettes.

\subsection{Detection model}
In our proposed framework, various kinds of object detection models can be adopted either for the coarse detection module or the fine detection module, and it's not limited to one certain object detection model. In this paper, we utilized the basic object detection model YOLOv5\cite{glenn_jocher_2022_6222936} and faster rcnn\cite{ren2015faster-rcnn} to construct the coarse detection model and fine detection model for our ablation study experiments.
\begin{figure*}%[t]
  \centering
   \includegraphics[width=0.8\linewidth]{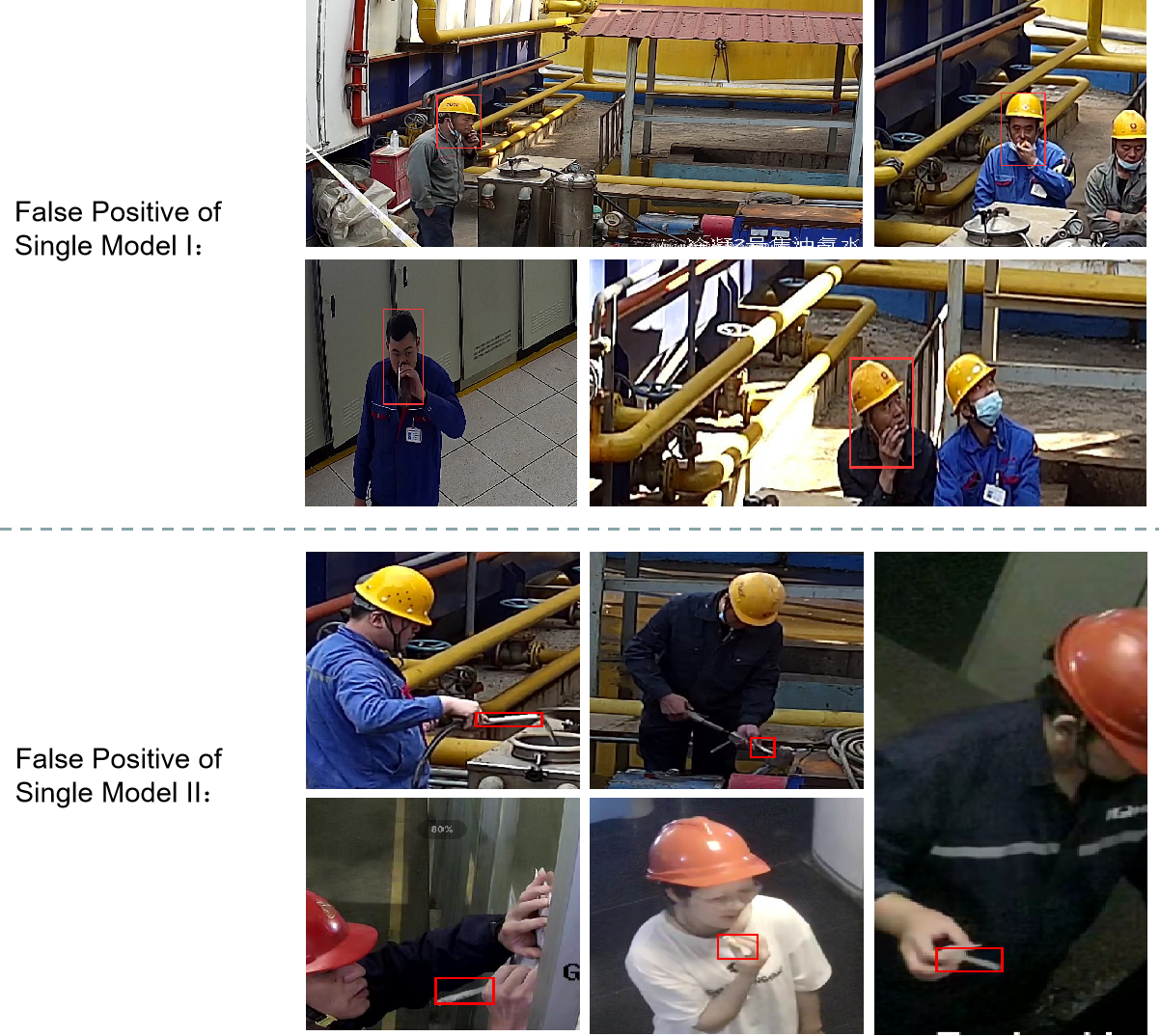}
   \caption{Examples of images containing false positive target. Single model I and single model II both make mistakes as a result of its intrinsic shortages.}
   \label{fig:falsePositive}
\end{figure*}
\section{Experiments}
In this section, we will firstly introduce our dataset, which we collected from all kinds of real-world scenarios, inlcuding chemical plants, industrial workshops and petrol stations, etc. Then we will describe the details of our experiments.

\subsection{Dataset}
\subsubsection{Coarse detection model's dataset}
We collected images from public webs, manual simulation and certain real-world scenarios, inlcuding chemical plants, industrial workshops and petrol stations, etc. A glimpse of our dataset is shown as Figure \ref{fig:original-images}. These images cover various gestures, angles, facial features, ages and illuminations of smoking cigarettes, which can enhance the detection models' genaralization ability while simultaneously maintaining a rather high accuracy. The smoking poses in these images are annotated by the way as shown in Figure\ref{fig:roughModle}.
\subsubsection{Fine detection model's dataset}
Based on the coarse detection model's dataset, the fine detection model's dataset is constructed by applying the coarse object detection and fine annotation. A glimpse of the fine detection model's dataset is shown as Figure \ref{fig:normalbbox}. And cigarettes are annotated by the way as shown in Figure \ref{fig:finerModle}.
\subsubsection{Additional dataset}
Smoking cigarettes, as a safety event which requires intense and precise attention and monitoring, sometimes
requires high precision under some circumstances while sometimes requires high recall under other circumstances. In view of this realistic dilemma, we additionally collected images without cigarettes inside but are very likely to be classified as smoking cigarettes. In this way, the dataset can enhance the models' generalization ability to certain degree, which will increase precision and recall simultaneously.
\subsection{Experimental settings}
The coarse detection model's input size is 1280x1280, and batch size is 64, and we train the model on a 8*V100 machine.
In view of the fine detection model's target is detected based on the coarse model's output, which most of the time
is rather small, we set the fine model's input size as 320x320, and batch size 100, and we train the model on a 4*2080Ti machine.
For both models, we use warmup epochs as 2, momentum for warmup epochs is 0.5, learning rate is set as 0.0032, and
momentum is 0.843.
\subsection{Experimental results}
In this section, we conduct experiments on our collected dataset. Both the coarse model and fine model are trained using yolov5m and faster rcnn. Additionally, to compare with our coarse-to-fine models, the single model focusing on human smoking pose and the one focusing on cigarettes are trained using yolov5m and faster rcnn.
% \begin{table}[htbp]
%   \centering
% 	\caption{Accuracy on test data of Yolov5. Single model I focuses its attention on the overall human body and smoking pose, while single model II focuses more attention on th ecigarette itself.}
%   % \resizebox{1\textwidth}{!}{
% 		\begin{tabular}{lc}
% 			\hline
% 			 models &accuracy  \\
% 			 \hline
% 				single model I & 0.714\\
%         single model II & 0.753\\
% 				coarse-to-fine model& 0.921   \\
% 			% xxxx & ZZZZ \\
% 				\hline
% 			\end{tabular}
% 		% }
%     \label{table:accuracy}
% \end{table}
\begin{table}[]
  \centering
  \caption{Accuracy on test data for different frameworks. Single Model I focuses on the overall human smoking pose, Single Model II focuses more on the cigarette itself, and Coarse-to-fine Models is our proposed framework.}
  \begin{tabular}{lll}\hline
  \pmb{Models}                   & \pmb{Frameworks}               & \pmb{Accuracy} \\ \hline
  \multirow{3}{*}{Yolov5}      & Single Model I       & 0.714    \\
                               & Single Model II      & 0.753    \\
                               & \pmb{Coarse-to-fine Models} & \pmb{0.921}   \\\hline
  \multirow{3}{*}{Faster RCNN} & Single Model I       & 0.704    \\
                               & Single Model II      & 0.734    \\
                               & \pmb{Coarse-to-fine Models} & \pmb{0.913} \\ \hline
  \end{tabular}
  \label{table:accuracy}
\end{table}

After training the models, we conduct ablation study on the models to show the superiority of our framework. We manually choose 450 positive images containing positive targets, and 400 negative images which contain no positive targets but are very likely to be detected as smoking cigarettes.

The results of YOLOv5 and faster rcnn are shown as Tabel \ref{table:accuracy}, Single Model I focuses on the overall human smoking pose, while Single Model II focuses more on th ecigarette itself. The results of both YOLOv5 and faster rcnn have the same regular pattern. In detail, for the 450 positive images, Single Model I, Single Model II and our Coarse-to-fine Models both achieve satisfactory results; as for the other 400 negative images, however, the two single models both detect many targets as smoking cigarettes, while the images have no smoke at all, which makes much more mistakes than our coarse-to-fine model.

In Figure \ref{fig:falsePositive}, we show several typical mistakes of detecting images containing false positive targets as smoking cigarettes. For instance, take the top left image in Figure \ref{fig:falsePositive} as an example, Single Model I detects the target as a true positive target, in which the person inside feeds nothing into the mouth but the action pose is similar to smoking cigarettes; however, in our method, the fine model correctly ignores this target and detects nothing as a true positive target, in other words, the person with the smoking pose but without cigarette is detected as a true negative target, which is classified corretly. In this way, a method utilizing one single model focusing on overall human smoking pose makes a mistake while our method does it precisely correctly.

Furthermore, take the top right image in Figure \ref{fig:falsePositive} as an example, Single Model II detects the target as a true positive target, in which the person inside is holding a pen only, the action pose is not similar to smoking cigarettes but the pen itself looks like a cigarette; however, in our method, the coarse model, which focuses on the overall human smoking pose, correctly ignores this target and detects nothing as a true positive target. In this way, a method utilizing one single model focusing on cigarette itself makes a mistake while our method does it precisely correctly.

\section{Conclusion}
In this paper, we propose a hierarchical object detection framework for hand-held action detection, which is composed of a coarse detection model to localize the human action pose and the fine detection model to identify the object iteself. Taking smoking cigarette detection for example, the dataset is collected from various practical application scenarios, and annotated in both coarse manner and fine manner. Based on typical basic models YOLOv5 and faster rcnn, the coarse model and fine model are trained, and compared with single model frameworks. The experimental results show that the coarse-to-fine framework achieves better results and effectively reduces false alarms than the single model frameworks. Our framework, as a new application-driven AI paradigm, can be further generalized to handle various kinds of hand-held action detection such as smoking, dialing and eating.

%%%%%%%%% REFERENCES
{\small
\bibliographystyle{ieee_fullname}
\bibliography{egbib}
}

\end{document}